\documentclass[sigconf]{acmart}
\AtBeginDocument{%
  \providecommand\BibTeX{{%
    \normalfont B\kern-0.5em{\scshape i\kern-0.25em b}\kern-0.8em\TeX}}}

\copyrightyear{2023}
\acmYear{2023}
\setcopyright{rightsretained}
\acmConference[GECCO '23 Companion]{Genetic and Evolutionary Computation Conference Companion}{July 15--19, 2023}{Lisbon, Portugal}
\acmBooktitle{Genetic and Evolutionary Computation Conference Companion (GECCO '23 Companion), July 15--19, 2023, Lisbon, Portugal}\acmDOI{10.1145/3583133.3590729}
\acmISBN{979-8-4007-0120-7/23/07}

\begin{document}

\title{Generative Meta-Learning Robust Quality-Diversity Portfolio}


%\author{Anonymous Author}
%\affiliation{
%  \institution{Anonymous Institution}
%  \city{City}
%  \country{Country}}
%\email{author@anonymous.com}

\author{Kamer Ali Yuksel}
\affiliation{
  \institution{aiXplain, inc. (AI Labs)}
  \streetaddress{16535 Grant Bishop Lane}
  \city{Los Gatos, CA 95032}
  \country{US}}
\email{kamer@aixplain.com}

\begin{abstract}
This paper proposes a novel meta-learning approach to optimize a robust portfolio ensemble. The method uses a deep generative model to generate diverse and high-quality sub-portfolios combined to form the ensemble portfolio. The generative model consists of a convolutional layer, a stateful LSTM module, and a dense network. During training, the model takes a randomly sampled batch of Gaussian noise and outputs a population of solutions, which are then evaluated using the objective function of the problem. The weights of the model are updated using a gradient-based optimizer. The convolutional layer transforms the noise into a desired distribution in latent space, while the LSTM module adds dependence between generations. The dense network decodes the population of solutions. The proposed method balances maximizing the performance of the sub-portfolios with minimizing their maximum correlation, resulting in a robust ensemble portfolio against systematic shocks. The approach was effective in experiments where stochastic rewards were present. Moreover, the results (Fig. 1) demonstrated that the ensemble portfolio obtained by taking the average of the generated sub-portfolio weights was robust and generalized well. The proposed method can be applied to problems where  diversity is desired among co-optimized solutions for a robust ensemble. The source-codes and the dataset are in the supplementary material.
\end{abstract}

\begin{CCSXML}
<ccs2012>
<concept>
<concept_id>10010147.10010257.10010321.10010333.10010334</concept_id>
<concept_desc>Computing methodologies~Bagging</concept_desc>
<concept_significance>500</concept_significance>
</concept>
<concept>
<concept_id>10010147.10010257.10010293.10011809.10011815</concept_id>
<concept_desc>Computing methodologies~Generative and developmental approaches</concept_desc>
<concept_significance>500</concept_significance>
</concept>
<concept>
<concept_id>10010147.10010257.10010321.10010333</concept_id>
<concept_desc>Computing methodologies~Ensemble methods</concept_desc>
<concept_significance>500</concept_significance>
</concept>
<concept>
<concept_id>10003752.10003809.10003716.10011138.10011140</concept_id>
<concept_desc>Theory of computation~Nonconvex optimization</concept_desc>
<concept_significance>500</concept_significance>
</concept>
<concept>
<concept_id>10003752.10003809.10003716.10011138.10011803</concept_id>
<concept_desc>Theory of computation~Bio-inspired optimization</concept_desc>
<concept_significance>500</concept_significance>
</concept>
<concept>
<concept_id>10010147.10010257.10010293.10011809</concept_id>
<concept_desc>Computing methodologies~Bio-inspired approaches</concept_desc>
<concept_significance>500</concept_significance>
</concept>
</ccs2012>
\end{CCSXML}
\ccsdesc[500]{Computing methodologies~Ensemble methods}
\ccsdesc[500]{Computing methodologies~Bagging}
\ccsdesc[500]{Theory of computation~Nonconvex optimization}
\ccsdesc[500]{Theory of computation~Bio-inspired optimization}
\ccsdesc[500]{Computing methodologies~Bio-inspired approaches}

%\keywords{robust portfolio optimization, deep generative models, out-of-sample generalization, meta-learning, ensemble learning, quality-diversity}
\keywords{robust portfolio optimization, ensemble learning, quality-diversity}

\begin{teaserfigure}
  \includegraphics[width=\textwidth]{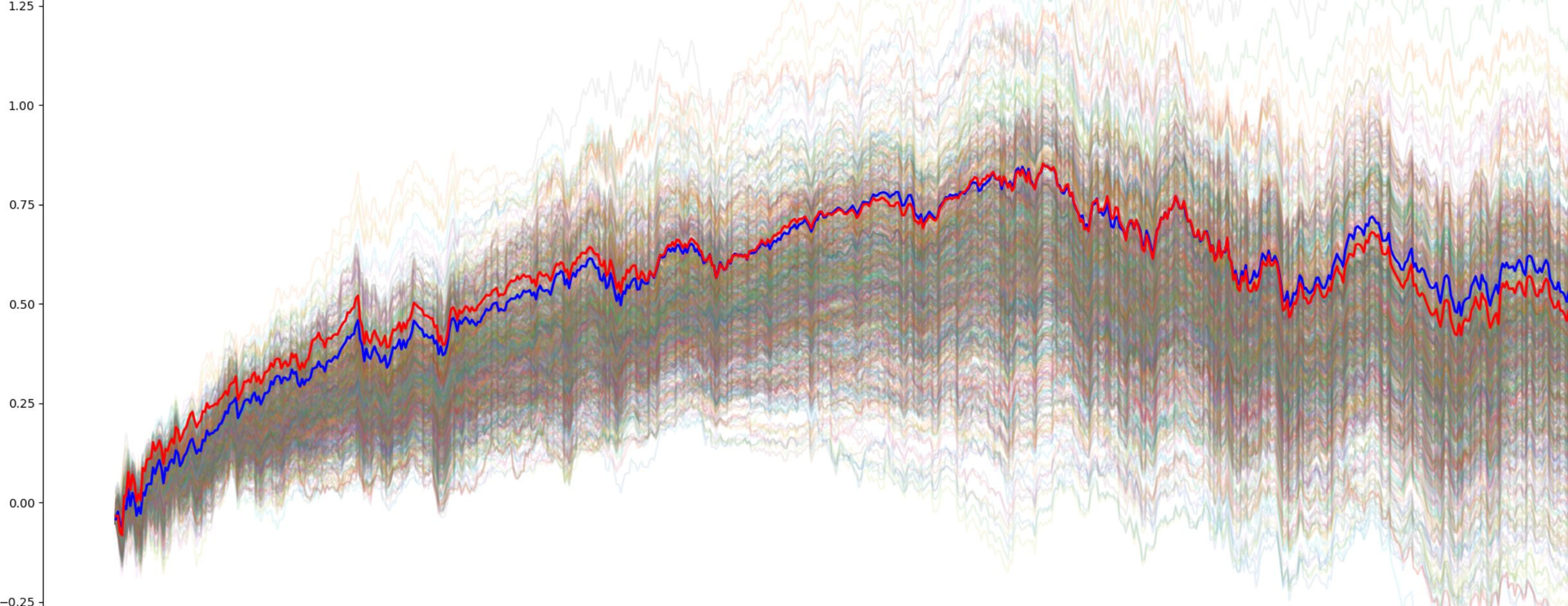}
  \caption{The index to be tracked (red), the sparse ensemble portfolio (blue), and behaviorally-diverse sub-portfolios co-optimized}
  \label{fig:teaser}
\end{teaserfigure}

%\received{20 February 2023}
%\received[accepted]{5 June 2023}

\maketitle

\section{Introduction}
Portfolio optimization is a fundamental problem in finance and has been widely studied. The traditional approach to portfolio optimization is based on quadratic or convex optimization, which assumes that the returns of the assets are normally distributed and have a constant covariance matrix \cite{markowitz}. This assumption is often unsatisfactory in practice and may lead to suboptimal solutions \cite{de2016building}. Moreover, quadratic optimization becomes computationally expensive in the case of large-scale portfolios. However, the portfolio optimization problem is typically non-convex, high-dimensional, and complex, which makes it challenging to find the global optimum. Over the years, various optimization methods and models have been proposed to address this challenge. The related works in this area have largely focused on optimizing portfolios using gradient-based methods, such as stochastic gradient descent (SGD), or heuristics, such as genetic algorithms. More recent techniques, such as evolutionary algorithms and particle swarm optimization, have been proposed to address the non-convex and high-dimensional optimization problems. These methods are based on generating a population of candidate solutions and optimizing them through an iterative process. However, these methods can be computationally expensive and may struggle to find a global optimum in large-scale non-convex optimization problems with noisy or stochastic rewards \cite{schaul2012benchmarking, li2013benchmark}.

Another important aspect of portfolio optimization is out-of-sample generalization. The portfolio performance should not be evaluated based on the training data but also on new, unseen data. Robust portfolio optimization is a crucial problem in finance, as it aims to optimize an investment portfolio while considering various sources of uncertainty and risk. The traditional approach to portfolio optimization is mean-variance optimization (MVO), which assumes that the returns of the assets are normally distributed \cite{markowitz}. However, this approach can be unreliable in real-world situations, as financial markets are often characterized by extreme events, such as sudden drops or spikes in asset prices. To address this problem, robust portfolio optimization techniques have been developed to account for the presence of outliers and other sources of risk. One of the most widely used approaches to robust portfolio optimization is the concept of diversification or risk parity. The portfolio should be diversified to minimize the risk, which can be achieved by having a multi-objective function. The risk-parity refers to holding a mix of different assets in the portfolio to reduce the overall risk and having allocation based on their volatility. Monte-Carlo simulation is often used for measuring out-of-sample robustness of a portfolio \cite{lopez2019tactical}. Stochastic optimization methods, such as Monte-Carlo optimization, can consider the uncertainty in the returns of the assets to ensure that the portfolio performance remains stable despite them.

A common approach to robustness in machine-learning is bagging \cite{breiman1996bagging}, a type of ensembling where several models are trained on the randomly selected subsets of the data, and their predictions are combined to form a single prediction. This approach helps reduce the risk of overfitting the training data and improves the model's overall accuracy, and stability \cite{fort2019deep}. By having diverse models, the weaknesses or biases of a single model can be compensated by the strengths of others, leading to improved overall performance and robustness. In robust portfolio optimization, bagging can be applied by training multiple portfolios over different parts of historical returns and then combining them to form a single ensemble portfolio. The goal is to diversify the risk into many uncorrelated bets so that the failure of the minority of them is less likely to impact the overall performance of the ensemble negatively. The use of bagging can also help reduce the risk of overfitting to a specific historical market condition, which would be crucial in achieving out-of-sample generalization and robustness in portfolio optimization.

Deep learning models, such as neural networks, can model complex relationships and achieve high accuracy in various tasks. In this context, deep generative models can be applied to the portfolio optimization problem, for meta-learning to generate high-performing and diverse portfolios. This work proposes a meta-learning approach for large-scale non-convex optimization of a robust portfolio ensemble. The approach uses a deep generative model to meta-learn the optimization process, to find high-quality, behaviorally diverse sparse sub-portfolios. The proposed method achieves superior out-of-distribution robustness and generalization compared to traditional optimization methods. The approach is also suitable for problems where stochastic (or noisy) rewards are present, and can be applied to robust optimization or ensemble learning (bagging), where in-population diversity (including behavioral) is desired.

The contributions of this paper are four-fold: (1) a novel meta-learning approach using a deep generative model is proposed for optimizing a robust portfolio ensemble; (2) the method balances maximizing the performance of the sub-portfolios with minimizing their maximum correlation, resulting in a robust ensemble portfolio against systematic shocks; (3) experiments demonstrate that the ensemble portfolio obtained by bagging (taking the average of) generated sub-portfolio weights is robust and generalized well; (4) the method can be applied to problems where behavioral diversity is desired among co-optimized solutions for a robust ensemble.

\section{Background}

Diversification is one of the fundamental concepts in portfolio optimization and has been the subject of numerous academic studies over the years. It has been long acknowledged that diversification is the key to reducing risk in an investment portfolio. The idea behind diversification is simple - spreading investment across multiple assets reduces the impact of market fluctuations on the portfolio's performance. However, traditional diversification methods based on per-asset risk diversification metrics (e.g., Max. risk contribution, Diversification ratio, Concentration ratio) have been criticized for being inadequate in capturing the complexities of the financial market. These metrics do not fully consider the inter-asset correlations and hence, are vulnerable to systemic shocks that can affect all assets in a portfolio.
Various risk-parity methods have been proposed in the literature to address this issue. Hierarchical Risk Parity (HRP) aims to construct a portfolio that can be decomposed into uncorrelated bets, thus reducing the exposure to both idiosyncratic and systemic shocks. The HRP method involves constructing a hierarchy of assets and allocating risk to each level in a balanced manner \cite{de2016building}. This results in a more robust portfolio that is less vulnerable to market fluctuations. Another method is Eigen-portfolios \cite{meucci2009managing}, based on the eigenvalue decomposition of the covariance matrix of the assets. The eigenvalue decomposition can be used to extract the underlying sources of risk in the market and allocate risk in a balanced manner among assets that are uncorrelated to each other. Eigen-portfolios effectively capture the financial market's underlying structure and provide a more robust portfolio.

Bagging, or bootstrapped aggregating, is an ensembling technique in machine learning that trains multiple models on different bootstrapped training data samples. The final prediction is then made by aggregating the predictions of the individual models, typically through a majority vote or an average. These techniques effectively improve the out-of-distribution robustness generalization performance of machine-learning models \cite{breiman1996bagging, fort2019deep}. This work aimed to transfer this ensembling technique to portfolio optimization; while also taking advantage of Quality-Diversity (QD) optimization \cite{pugh2016quality, cully2017quality}, a population-based multi-objective optimization technique that aims to find a diverse set of high-quality individual solutions rather than one. The assumption was as follows: an ensemble portfolio from many high-quality, behaviorally diverse sub-portfolios, would improve the out-of-sample generalization and robustness. However, the global optimization of such a multi-objective portfolio ensemble was challenging due to multiple local optima and the difficulty of finding the global optimum. In recent years, population-based optimization has been proposed as a promising approach to tackle non-convex optimization problems. The population-based optimization methods such as Evolutionary Strategies \cite{sun2009efficient, wierstra2014natural}, search the solution space using a population of candidate solutions, inspired by biological evolution. QD optimization (illumination) is a promising non-convex optimization approach emphasizing maintaining population diversity while exploring the solution space.

\section{Methodology}

The proposed method in this paper aims to achieve a robust portfolio ensemble through a meta-learning approach. This approach aims to generate diverse and high-quality sub-portfolios that can be combined to form a single portfolio. The method utilizes a deep generative model to achieve this objective, which consists of a convolutional layer, a stateful LSTM module, and a dense network. The convolutional layer transforms the input noise into a desired distribution in latent space, while the LSTM module adds dependence between generations. The dense network decodes the population of solutions from the latent space to the solution space. The deep generative model takes a randomly sampled batch of Gaussian noise as input and outputs a population of solutions to be evaluated using the objective function of the problem. The generative model parameters are updated at each iteration using the gradients obtained from the objective function, with AdamW optimizer \cite{loshchilov2017decoupled}, commonly used in deep learning. The objective function of this study focuses on index tracking, where the aim is to track a given index with sparse portfolio weights. Index tracking can be considered a generalized version of the minimum-variance portfolio where the returns of a given index are used instead of a target return when calculating the deviations. The sparsity is achieved without L1 regularization \cite{tibshirani1996regression} by applying softmax activation during training, and sparsemax activation \cite{martins2016softmax} during the validation. The objective function for index tracking was the mean squared error in-between the daily returns of each sub-portfolios and the given index. To achieve robustness, the proposed method also tries to minimize the maximum correlation in-between sub-portfolios, while minimizing the objective function for portfolio optimization, resulting in a robust ensemble portfolio against systematic shocks. The behavioral diversity measured from the correlation matrix of sub-portfolio returns is preferred against diversity based on weights to increase the robustness against systematic shocks \cite{de2016building, meucci2009managing}. To further increase the robustness of the ensemble portfolio, stochastic optimization techniques are also applied, such as randomly corrupting the candidate portfolio weights by adding noise or zeroing them. This aimed to obtain portfolio candidates that are not depending their performance on specific assets. Furthermore, a monte-carlo optimization technique has been performed by calculating the candidate rewards of the sub-portfolios on randomly selected historical periods of return to prevent them from overfitting to a specific historical period.

The proposed method has many advantages over popular convex optimization methods and heuristic algorithms for non-convex optimization. The deep generative model allows for meta-learning the optimization process, enabling it to find high-quality, behaviorally diverse sparse sub-portfolios. This results in a robust ensemble portfolio that has improved out-of-distribution robustness and generalization. Additionally, the method can handle stochastic rewards and be applied to problems where behavioral diversity is desired among co-optimized solutions. The results (Table 1) demonstrated that the ensemble portfolio obtained by taking the average of the generated sub-portfolio weights was robust and generalized well. This meta-learning approach is suitable for problems where behavioral diversity is desired among co-optimized solutions and can be applied to a wide range of portfolio optimization problems, including index tracking. In conclusion, the proposed method combines the strengths of deep learning and ensemble learning (bagging) to achieve a robust portfolio ensemble through meta-learning. Using a deep generative model and the optimization approach provides an efficient and effective solution to the portfolio optimization problem, particularly when the objective is sparse index-tracking.

\begin{table}
  \caption{The sparse index tracking errors achieved for various ETF(s) by different optimizers, in the out-of-sample period.}
  \label{tab:optimizer}
  \begin{tabular}{cccc}
    \toprule
    Optimizer & RTH & RYH & VGT \\
    \midrule
    SGD & 0.000087 & 0.000091 & 0.000175 \\
    CMA-ES & 0.000039 & 0.000030 & 0.000068 \\
    RAdam & 0.000032 & 0.000025 & 0.000065 \\
    NAdam & 0.000042 & 0.000031 & 0.000040 \\
    RMSprop & 0.000028 & 0.000015 & 0.000037 \\
    Adagrad & 0.000024 & 0.000009 & 0.000031 \\
    Adam & 0.000023 & 0.000011 & 0.000027 \\
    AdamW & 0.000023 & 0.000011 & 0.000027 \\
    Adamax & 0.000021 & 0.000009 & 0.000027 \\
    Rprop & 0.000012 & 0.000009 & 0.000015 \\
    \textbf{Proposed} & \textbf{0.000010} & \textbf{0.000008} & \textbf{0.000011} \\
    \bottomrule
  \end{tabular}
\end{table}

\section{Experiments}

The performance of the proposed generative meta-learning method was evaluated through experiments conducted on a large-scale sparse index tracking problem. The validation performance was compared with that of popular optimizers in deep learning. The experiments were performed on a regular desktop computer with a NVIDIA GeForce RTX3090 GPU and each optimizer was run for fifty training iterations. The training loss function of the proposed method was stochastic and included a secondary objective of minimizing the maximum correlation among the population, as opposed to the loss function utilized for the optimizers during training.

A real-world US stock-market dataset was utilized to test the proposed meta-learning approach for robust portfolio ensemble optimization. The experiments were carried out using market data for 2458 stocks over 14 years right after 2007-2008 financial crisis, from 2009 to 2023, with 80\% of data until the beginning of 2020 being kept for out-of-sample validation. Due to the several black swan events in-between 2020 and 2023, such as the Covid-19 pandemic and the Russian Invasion of Ukraine, the resulting validation set can be considered as out-of-distribution \cite{lipton2020three}, compared to the training period that was a consistent bull-market with no major events. This made the selected validation set a perfect playground for comparing the robustness and generalization of the optimization methods.

The proposed method was compared with SGD, CMA-ES, RAdam, NAdam, RMSprop, Adagrad, Adam, AdamW, Adamax, and Rprop optimization algorithms based on the sparse index tracking error, which measures the L2-norm of deviations between the logarithmic returns of the target index and the sparse portfolio. The same number of maximum function evaluations were used for each optimizer, and their best validation performance achieved over the training iterations was reported. As demonstrated in Table 1, the validation performance of the proposed method was superior to that of the other optimizers in sparse tracking of Retail, Healthcare, and Information-Technology indexes (referred to as RTH, RYH, and VGT, respectively in Table 1). The dataset and source codes used in the experiments are available in the supplementary material.

The results of the experiments revealed that the proposed method had the best sparse index tracking error among the other optimizers during the out-of-sample validation period. The effectiveness of the proposed method in optimizing a robust ensemble portfolio in the presence of stochastic rewards was demonstrated. Furthermore, the results showed that the ensemble portfolio obtained through the averaged sub-portfolio weights achieves better out-of-distribution robustness and generalization. In conclusion, the proposed meta-learning approach for robust portfolio ensemble optimization was found to be promising through the experiments conducted. The proposed method outperformed popular optimization algorithms regarding sparse index tracking error, showing the effectiveness of using a deep generative model to generate diverse and high-quality sub-portfolios for the ensemble. The results indicated that the proposed method was robust and generalized well, making it a potential solution for other robust optimization problems.

\section{Discussion}

Experimental results demonstrated the efficacy of using a meta-learning approach to generate a robust portfolio ensemble in the presence of stochastic rewards. The proposed method leverages a deep generative model to generate diverse and high-quality sub-portfolios combined to form the ensemble portfolio. The results showed that the ensemble portfolio obtained by taking the average of the generated sub-portfolio weights was robust and generalized well. A potential area of future-work is experimenting with more advanced methods for bagging the co-optimized sub-portfolios. In this work, sub-portfolios are simply averaged to form the ensemble portfolio, but more sophisticated methods can further improve the ensemble's validation performance. For instance, correlations between the sub-portfolios or their individual risks (variances) can be used during bagging to obtain a more robust ensemble portfolio. Furthermore, the ensemble's uncorrelated constituents can be ensembled dynamically during the out-of-sample validation (or while trading in the real-world application) using nowcasting \cite{lipton2020three} via the short-term observations continuously made on them.

\section{Conclusion}

In this work, a novel meta-learning approach has been proposed for optimizing a robust ensemble portfolio. The approach used a deep generative model to generate high-quality and behaviorally diverse sub-portfolios combined to form an ensemble portfolio. The generative model was trained with a gradient-based optimizer to maximize the performance of sub-portfolios and minimize the maximum correlation between their returns. The results of the experiments showed that the proposed method outperforms popular optimization methods in terms of out-of-distribution robustness and generalization. The proposed method effectively balances performance and diversification, ensuring that the ensemble portfolio is robust against systematic shocks. This demonstrates the potential of meta-learning quality-diversity optimization using deep generative models in robust portfolio optimization and opens the door for further exploration in other domains where robust ensemble learning is desired. The proposed method can be applied to problems where diversity is desired among co-optimized solutions, and where stochastic or noisy rewards are present. The use of deep generative models in meta-learning quality-diversity optimization is novel and provides a promising direction for future research. Future-research could extend the proposed method to more complex portfolio optimization problems and explore its application to other domains where robustness and generalization are desired. 

\bibliographystyle{ACM-Reference-Format}
\bibliography{sample-base}

\end{document}